\documentclass[conference]{IEEEtran}
\IEEEoverridecommandlockouts
\usepackage{cite}
\usepackage{amsmath,amssymb,amsfonts}
\usepackage{algorithmic}
\usepackage{graphicx}
\usepackage{subfig}
\usepackage{textcomp}
\usepackage{xcolor}
\def\BibTeX{{\rm B\kern-.05em{\sc i\kern-.025em b}\kern-.08em
    T\kern-.1667em\lower.7ex\hbox{E}\kern-.125emX}}
\begin{document}

\title{Autonomous Navigation System from Simultaneous Localization and Mapping \\}
\author{\IEEEauthorblockN{Micheal Caracciolo}
\IEEEauthorblockA{\textit{Electrical and Computer Engineering} \\
\textit{Clarkson University}\\
Potsdam, NY \\
caraccmv@clarkson.edu}
\and
\IEEEauthorblockN{Owen Casciotti}
\IEEEauthorblockA{\textit{Electrical and Computer Engineering} \\
\textit{Clarkson University}\\
Potsdam, NY \\
cascioo@clarkson.edu}
\and
\IEEEauthorblockN{Christopher Lloyd}
\IEEEauthorblockA{\textit{Electrical and Computer Engineering} \\
\textit{Clarkson University}\\
Potsdam, NY \\
lloydcd@clarkson.edu}
\and
\IEEEauthorblockN{Ernesto Sola-Thomas}
\IEEEauthorblockA{\textit{Electrical and Computer Engineering} \\
\textit{Clarkson University}\\
Potsdam, NY \\
schumae@clarkson.edu}
\and
\IEEEauthorblockN{Matthew Weaver}
\IEEEauthorblockA{\textit{Electrical and Computer Engineering} \\
\textit{Clarkson University}\\
Potsdam, NY \\
weavermd@clarkson.edu}
\and
\IEEEauthorblockN{Kyle Bielby}
\IEEEauthorblockA{\textit{Electrical and Computer Engineering} \\
\textit{Clarkson University}\\
Potsdam, NY \\
bielbyka@clarkson.edu}
\and
\IEEEauthorblockN{Md Abdul Baset Sarker}
\IEEEauthorblockA{\textit{Electrical and Computer Engineering} \\
\textit{Clarkson University}\\
Potsdam, NY \\
sarkerm@clarkson.edu}
\and
\IEEEauthorblockN{Masudul H. Imtiaz}
\IEEEauthorblockA{\textit{Electrical and Computer Engineering} \\
\textit{Clarkson University}\\
Potsdam, NY \\
mimtiaz@clarkson.edu}
}
\maketitle
\begin{abstract}
This paper presents the development of a Simultaneous Localization and Mapping (SLAM) based Autonomous Navigation system. The motivation for this study was to find a solution for navigating interior spaces autonomously. Interior navigation is challenging as it can be forever evolving. Solving this issue is necessary for a multitude of services, like cleaning, the health industry, and in manufacturing industries. The focus of this paper is the description of the SLAM-based software architecture developed for this proposed autonomous system. A potential application of this system, oriented to a smart wheelchair, was evaluated. Current interior navigation solutions require some sort of guiding line, like a black line on the floor. With this proposed solution, interiors do not require renovation to accommodate this solution. The source code of this application has been made open source so that it could be re-purposed for a similar application. Also, this open-source project is envisioned to be improved by the broad open-source community upon past its current state. 
\end{abstract}


\section{Introduction}
According to Precedence Research, the Autonomous vehicle market is expected to have a compound annual growth rate of 63.5\% from 2020 to 2027 [1]. This growing rate depends on the infrastructure for this autonomy also being improved as the technology grows. Certain applications of this autonomous navigation can easily be created. Some examples of use are a moving platform in a manufacturing warehouse, a gurney in a hospital, or even a robot delivering food in a hotel. Current solutions primarily have devices that have many sensors for the devices to navigate around an area autonomously  [2,3]. Some solutions can be described to be basic, with the use of Automated Guided Vehicles (AGV) that follow lines on the ground [4]. With our proposed solution, the device can require less sensors, is more flexible and provides a stronger interface between person and device if required. It is accomplished by making a map of the area with a singular camera, then having a device be tailored to take a path from start and end points, and to follow it with direct object avoidance from a Deep Learning based system.   
This paper talks about the software behind this approach, along with an example of using it. This implementation is based on this approach to autonomously navigate a wheelchair through a medical building, which is described later on. The paper is organized as follows: Section II provides the broad overview of the system; Section III discusses the system components; Section IV shows the system implementation and Section V provides conclusions of the study. All of the project’s design files have been made available in GitHub here: https://github.com/michealcarac/VSLAM-Mapping  
\begin{figure}[htbp]
\centerline{\includegraphics[width=0.4\textwidth]{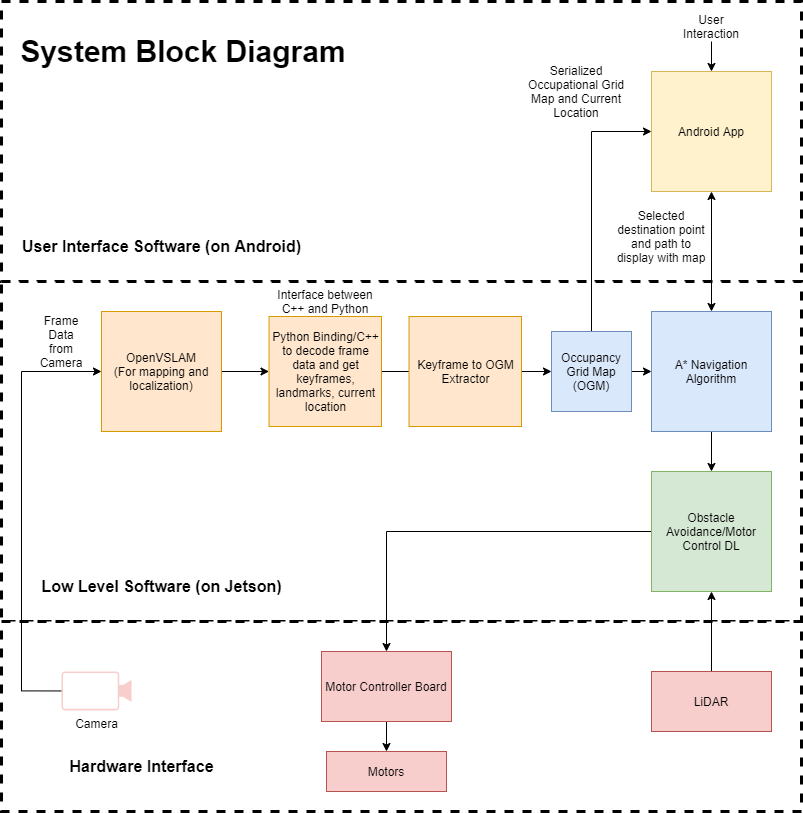}}
\caption{System Block Diagram}
\label{fig1}
\end{figure}

\section{System Overview}

The proposed approach to create this Autonomous Navigation System is made up of different sections. The first design task of this development was the selection of the camera system as most Simultaneous Localization and Mapping (SLAM) based programs use certain types of cameras. In some cases, 360 degrees, fisheye, stereoscopic, or monoscopic cameras are used [2,3]. After deciding a camera system, a form of SLAM (Simultaneous Localization and Mapping) will be chosen. SLAM is a camera based system that records frames in real-time and creates a virtual map of current, and previous positions. It also records the walls and other objects then plots them in the map, as shown by ORB-SLAM [5].This paper will talk about the use of Open Visual Simultaneous Localization and Mapping (OpenVSLAM) [6], which is a direct derivative of ORB-SLAM. Once a form of SLAM is chosen and used, a map file will be generated. A map file is a file generated by a SLAM program that has all of the data for the generated map. This includes keyframes (Position of the device as it records frames for SLAM), landmarks (Objects/walls detected while the device records frames for SLAM), and camera settings (From camera calibration, and the resolution/frames per second). This map file may come in a few different formats. OpenVSLAM exports in a “.msg” format, which is described as a MessagePack. This format requires a script to be opened properly. Once it is opened and the keyframes and landmarks are extracted, it is time to move on to generating the Occupancy Grid Map (OGM). Once it is created as an OGM, a path algorithm is chosen. For this paper, A* (an informed search algorithm) was the chosen path algorithm as it can be created to focus on 90-degree angles instead of diagonals by changing the heuristic function, which is ideal for an OGM. This of course can be changed so that diagonal paths can be used. Usage of a path algorithm requires a start and end position. The starting point can be acquired manually or by automatically reading from a GPS coordinate system. The end destination point is recommended to be generated by the user using the device, i.e whoever is being driven. This point can be acquired by user input from an Android interface via some sort of cell phone device. These two points are then fed into the path algorithm along with the OGM, which then will send a guided path to the Deep Learning model to process and run any sort of motor or movable system. The Deep Learning model, hosted on an embedded processor, requires this path so that it can use a Light Detection and Ranging (LiDAR) sensor to avoid any sort of objects in its way, and then get back onto the path. 
\section{System Components}
\subsection{Camera System}
To use a camera as a sensor, it has to be calibrated. An example camera is the Intel Realsense D415 [7] which has multiple cameras, one for Red Green Blue (RGB), Depth, and Infrared (IR). After calibrating the RGB camera, it can be used for Monocular SLAM as it is used to view the difference in color on objects around a room. This camera system is used in VSLAM to map an area.

\subsection{VSLAM}
\begin{figure}[htbp]
\centerline{\includegraphics[width=0.3\textwidth]{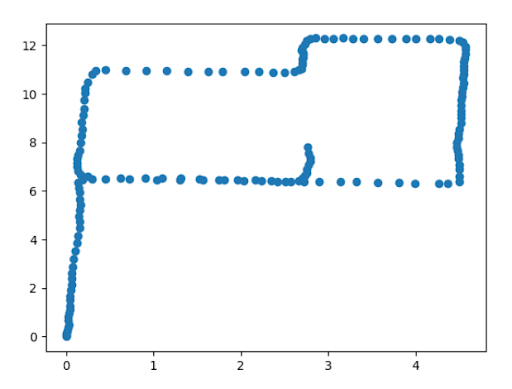}}
\caption{Plotted 2D Keyframe Points}
\label{fig2}
\end{figure}
In order to keep track of the real-world locations of the autonomous vehicle and the surrounding area, VSLAM [8] is used. VSLAM algorithms use frame data from an image sensor to identify key points, landmarks, and edge data. An example of these keypoints is illustrated in [Fig. 2]. These important pixels in an image frame are used to both create a map of an area, while concurrently estimating the current position of the camera using the map. [Fig. 3a] shows an example generated map. This allows VSLAM systems to be used for autonomous control of robots and drones, etc. 
	One specific VSLAM implementation, which was used for this project is OpenVSLAM [6]. OpenVSLAM is an open-source C++ based visual SLAM framework which works with many image sensors. This includes Monocular (single camera), Stereo (multiple cameras), and RGBD (color and depth camera). It is built on the foundations of other VSLAM frameworks including ORB–SLAM [5], LSD–SLAM [9], and DSO [10]. Maps can be stored and loaded using a MessagePack database file to allow for reuse with multiple applications. These ‘.msg’ database files store keyframe data, which is a list of locations where OpenVSLAM considers the image data to be a key point. These points can be used to indicate a map of locations traveled due to the frequency of these keyframes. OpenVSLAM is also designed to be easily extendable for specific applications and systems. 
To take care of these files, a MessagePack unpacker has to be used. It will pull landmarks and keyframes from the file and transform the points from a rotational matrix into x, y, and z coordinates. 
	For this study, OpenVSLAM was used to create a map of the area to navigate, while providing real-time updates on the location of the vehicle. A D415 camera was used to provide image data for OpenVSLAM to process while creating the map and localizing. In order to integrate this in the autonomous wheelchair system, first, a map is created of the area. Once this map has been created, OpenVSLAM is only responsible for sending the current location to the main software program. This localized point provides feedback to the system on the device’s current location with respect to the created map. 
\subsection{Occupancy Grid Map}

\begin{figure}[htbp]
\centering
  \subfloat[]{\includegraphics[width=0.08\textwidth]{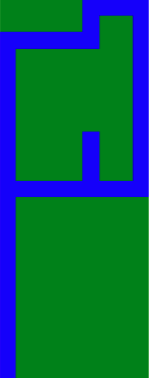}\label{fig:f1}}
  \hspace{60pt}
  \subfloat[]{\includegraphics[width=0.08\textwidth]{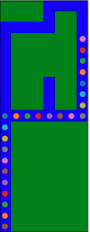}\label{fig:f2}}
  \caption{Converted Occupancy Grid Maps}
\label{fig3}
\end{figure}
The map data from VSLAM is converted to a grid map where the area of each square represents a real-world area. Each square is then given a value to indicate whether the square contains an obstacle or not. This custom grid map was based on the design of an occupancy grid map, which stores a float number for each tile corresponding to the probability that the tile is occupied or not. Occupancy grid maps are commonly used when dealing with sensor data with uncertainty. However, for this paper, each tile is only occupied or unoccupied which depends on if each tile contains a valid keyframe from OpenVSLAM or not. 
In order to create this grid map, the keyframes were parsed using Python from the “.msg” database file exported from OpenVSLAM. The keyframes were then transformed into a two-dimensional list from the three-dimensional keyframe data. A plot of these keyframes is illustrated in [Fig. 2]. Next, each keyframe point was used to mark the tile it was located in as unoccupied. This created a grid map that shows valid paths where the camera has been. An example map can be seen in [Fig. 3a], where occupied tiles are colored blue and unoccupied tiles are colored green. This map is used for navigation and display to the user as feedback of the surrounding area to navigate to.
\subsection{Point Selection}
An Android application is developed to allow user control over the autonomous navigation system. On startup, the app sends a JavaScript Object Notation (JSON) array request to the server. The JSON OGM object is then transmitted to the application by the server. The application then parses through the JSON object and creates a local map object to be used by the rest of the application. From this data, a grid layout of squares is displayed on the screen. If the JSON object value that is parsed is greater than or equal to 0.5, then the propagated square is determined as occupied and would have a determined occupied color. Otherwise, if the parsed OGM value is less than 0.5, the propagated square is determined as unoccupied and will be set to the predetermined color. This process is used to create the user control map on the Android device’s display.
	After the control map is created, the user then selects a square on the map to determine the final destination of the device. The relative coordinates of the selected square are then transmitted to the server over WebSocket. Once the server receives this information, the optimized path to the final destination is transmitted as a JSON array over the WebSocket to the client. As the autonomously navigated device traverses the predetermined path, the current position of the device is transmitted to the Android application and displays on the Android device by changing the color of the corresponding position square to the predetermined marker. This allows the user to be aware of the current location of the navigating device at all times.
\subsection{A*}
To ensure the obtained path is always the best path and is done automatically given a start and endpoint, a single A* algorithm is used. After mapping, a generated OGM is imported, where A* will start at a defined point and look at neighboring points until it eventually reaches the designated end position on the OGM. It then takes a heuristic function to make a decision on which path is the best option. In this case, it decides which path is the least detrimental i.e., the higher number of points traversed. The created path is used to drive any sort of mobile device from a start point to endpoint. This is illustrated in [Fig. 3b]. 
\subsection{LiDAR}
A LiDAR sensor is connected to the system to allow for a simpler view of the environment instead of having to do image processing twice, once for SLAM and once with the LiDAR. The LiDAR will be able to provide distance measurements in a range of directions. The specific LiDAR in the system is the YDLiDAR X2L. It is capable of 360 degrees ranging from up to 8 meters. This sensor is configured to only return a 90-degree section directly in front of the system for the most coverage without overloading the system with data. An example of this 90-degree section is shown [Fig. 4].
\subsection{Deep Learning Model}

\begin{figure}[htbp]
\centerline{\includegraphics[width=0.21\textwidth]{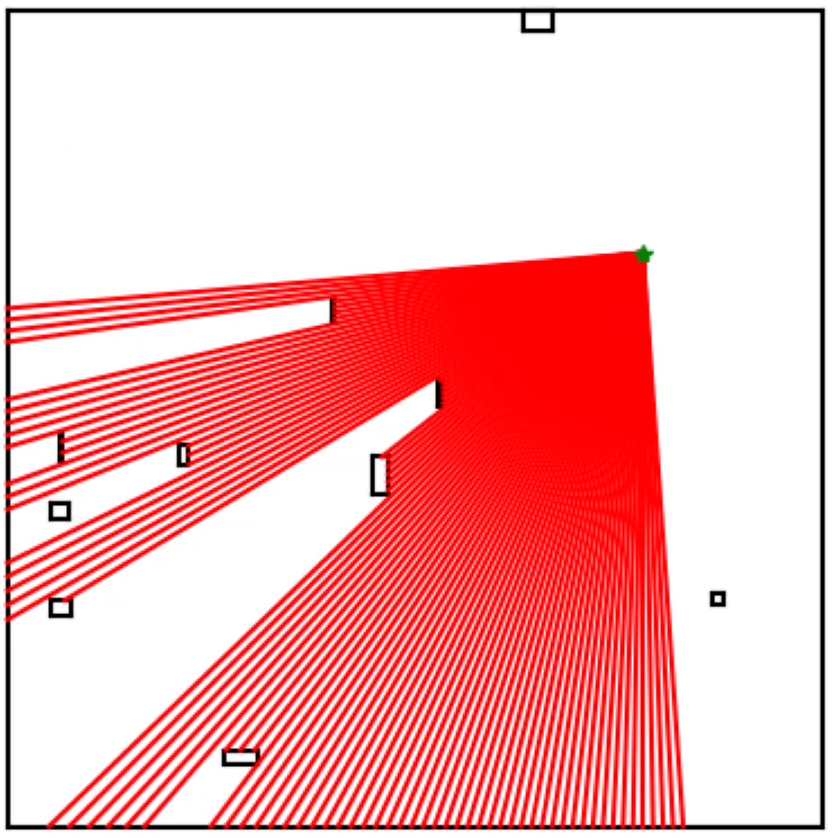}}
\caption{Ray Casting}
\label{fig4}
\end{figure}
This neural network (NN) is designed and based on the ideas of DeepMind [11] and their Deep Q Learning models. This takes in the data from the LiDAR sensor, the known position of the system, and the desired path to follow. The output from the model is the desired direction to move in that will be sent to the motor controllers. The structure of the model is a fully dense model that reduces the features down each layer until reaching a single feature that corresponds to the direction. The model trains on simulations of this environment in order to learn how to map these inputs into usable outputs to control the system. This training happens on a dedicated lab computer and the model inferencing is performed on the Jetson Xavier. Ideally, the use of A* to generate the path will reduce the complexity of the problem, giving the NN a better chance of converging to a solution. The goal is to have this model follow the given path and avoid any obstacles detected by the LiDAR.

\section{System Implementation}
This paper’s approach is used in a real-life situation to navigate an autonomous wheelchair throughout a building. This is done using A*, OpenVSLAM, and the Occupancy Grid Maps. It is tested in a closed environment lab, where a map was created. This lab is one big room with multiple large desks around. The wheelchair has slight issues with motor control as one motor moves a bit faster than the other. The wheelchair however, is able to accurately navigate different designated paths [Fig. 5] while using a SLAM mapped area. The Deep Learning model is not ready yet for this type of work, therefore it is not used in the example. This example ran for a few hours and shows promising results. It can accurately go from one side of the room to the other with a simple start and end value. The mechanical and hardware design of the wheelchair is described in more detail in the paper “Design of an Initial Prototype of the AI Wheelchair” [12] along with all of the files being publicly available on GitHub here: https://github.com/michealcarac/VSLAM-Mapping
\begin{figure}[htbp]
\centerline{\includegraphics[width=0.1208\textwidth]{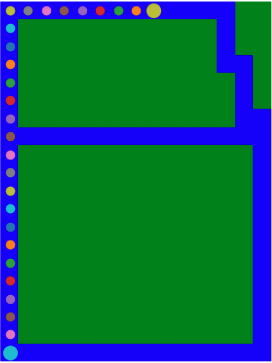}}
\caption{Wheelchair A* Path on OGM}
\label{fig5}
\end{figure}
\section{Conclusion and Discussion}
The results of this paper show a promising future for this technology. The simplicity of the design allows for greater flexibility of the system. Using open-source components and software means the community of talented and dedicated developers are able to push this design further as new techniques and approaches become available. While most of the system was able to come together, one major struggle that still continues is the use of Deep Learning to avoid obstacles. This turns out to be far more complicated than expected. A simple environment is set up to test the concepts of this model. The goal is to navigate inside a box to a point across the room without any obstacles in the way. The idea is to determine the amount of training that is needed for the larger and more complicated tasks in the future. Due to some undetermined reasons, the model is not able to learn to navigate this environment in any amount of training. The agent either spins for the allowed duration of the test or continues moving forward until crashing into the wall. One reason for this behavior might be the lack of normalization of the input data. In the environment, the LiDAR data is simulated by using rays cast from the agent to determine the distance to the nearest object. This is then passed directly into the model. It is uncertain if clipping these values and then normalizing them results in better performance. Overall, the paper’s technology fills a much-needed gap in the market of autonomous navigation systems by being simpler and more flexible than others. 

\end{document}